\documentclass[sigplan,10pt]{acmart}

\usepackage{amsmath}
\usepackage{booktabs}
\usepackage{enumitem}
\setlist{topsep=3pt, itemsep=1pt, parsep=0pt}
\usepackage{listings}
\usepackage{array}
\usepackage{multirow}
\usepackage{ragged2e}
\usepackage{xcolor}
\usepackage{url}
\usepackage{multicol}
\usepackage{tikz}
\usetikzlibrary{arrows.meta,positioning,fit,backgrounds,calc,decorations.pathreplacing}
\usepackage{pifont}
\newcommand{\cmark}{\ding{51}}

\copyrightyear{2026}
\acmYear{2026}
\setcopyright{cc}
\setcctype{by}
\acmConference[EuroMLSys '26]{Sixth European Workshop on Machine Learning and Systems}{April 27--30, 2026}{Edinburgh, Scotland Uk}
\acmBooktitle{Sixth European Workshop on Machine Learning and Systems (EuroMLSys '26), April 27--30, 2026, Edinburgh, Scotland Uk}
\acmDOI{10.1145/3805621.3807648}
\acmISBN{979-8-4007-2605-7/2026/04}
\usepackage{balance}

\newcolumntype{L}[1]{>{\RaggedRight\arraybackslash}p{#1}}

\lstdefinestyle{pseudocode}{
  basicstyle=\ttfamily\footnotesize,
  breaklines=true,
  columns=fullflexible,
  frame=single,
  showstringspaces=false,
  aboveskip=6pt,
  belowskip=4pt
}

\title{\textsc{ClawVM}: Harness-Managed Virtual Memory for Stateful Tool-Using LLM Agents}

\author{Mofasshara Rafique}
\affiliation{%
  \institution{Independent Researcher}
  \country{Switzerland}
}
\email{mofasshara@hotmail.com}

\author{Laurent Bindschaedler}
\affiliation{%
  \institution{Max Planck Institute for Software Systems}
  \city{Saarbr\"{u}cken, Saarland}
  \country{Germany}
}
\email{bindsch@mpi-sws.org}


\begin{document}

\begin{abstract}
Stateful tool-using LLM agents treat the context window as working memory, yet today's agent harnesses manage residency and durability as best-effort, causing recurring failures: lost state after compaction, bypassed flushes on reset, and destructive writeback.
We present \textsc{ClawVM}, a virtual memory layer that manages state as typed pages with minimum-fidelity invariants, multi-resolution representations under a token budget, and validated writeback at every lifecycle boundary.
Because the harness already assembles prompts, mediates tools, and observes lifecycle events, it is the natural enforcement point; placing the contract there makes residency and durability deterministic and auditable.
Across synthetic workloads, 12 real-session traces, and adversarial stress tests, \textsc{ClawVM} eliminates all policy-controllable faults whenever the minimum-fidelity set fits within the token budget, confirmed by an offline oracle, and adds median $<$\,50\,$\mu$s of policy-engine overhead per turn.
\end{abstract}

\maketitle

\section{Introduction}

General-purpose tool-using large language model (LLM) agents, persistent assistants that manage email, calendars, smart-home devices, messaging, web automation, and dozens of other services, now run for hours or days, accumulating state across hundreds of tool calls and multiple sessions.
OpenClaw~\cite{openclaw:context} and its derivatives (NanoClaw~\cite{nanoclaw}, PicoClaw~\cite{picoclaw}, ZeroClaw~\cite{zeroclaw}) are the most prominent examples; coding agents such as Claude Code, Codex CLI, and Cursor represent a narrower but equally demanding subclass.
These \emph{stateful tool-using agents} maintain durable state across sessions and operate over extended durations that span multiple context windows.

For these agents, the model context window is scarce working memory: it must hold constraints, the active plan, recent dialogue, and tool outputs needed for correct action, while transcripts and artifacts reside in durable backing stores.
Correct operation depends on having the needed state resident at the right time and fidelity.
When the needed state is absent, or the correct state is silently discarded, agents repeat work, violate user preferences, and lose progress mid-plan.
These are not rare edge cases: public issue trackers and practitioner reports document recurring rules and constraints lost after context summarization~\cite{openclaw:bootstrap-issues,openclaw:issue10524,openclaw:pr20267,practitioner:protocol-forgotten}, state silently dropped on session reset~\cite{openclaw:flush-bypass,openclaw:issue8185,openclaw:issue17034,openclaw:pr17041}, and persistence operations that overwrite rather than merge~\cite{openclaw:writeback-corrupt}.

Modern agent harnesses such as OpenClaw provide building blocks: pruning, retrieval, compaction, pre-compaction memory flushes, and external memory plugins~\cite{openclaw:context,openclaw:compaction,openclaw:memory,mem0:openclaw,cognee:openclaw}.
These improve recall quality, but none provides an enforceable contract over residency, durability, or auditability.
Practitioners have responded with configuration tweaks and add-on layers~\cite{openclaw:config,qmd,mem0:openclaw,cognee:openclaw,practitioner:compaction-fix}, yet flushes can still be bypassed, writeback can still be destructive, and no mechanism guarantees that critical state survives lifecycle transitions.
Operating systems learned this lesson decades ago: when a runtime manages a fast, scarce tier alongside slow, durable storage, the answer is virtual memory, not best-effort heuristics.
Prior work borrows the paging metaphor~\cite{memgpt2023,memos2025}, but no production harness enforces it.

\begin{figure*}[t]
\centering
\begin{tikzpicture}[
  >=Stealth,
  node distance=0.6cm and 0.8cm,
  box/.style={draw, rounded corners=2pt, minimum height=0.7cm, minimum width=1.8cm,
              font=\footnotesize\sffamily, align=center, thick},
  sbox/.style={box, minimum width=1.3cm, minimum height=0.55cm, font=\scriptsize\sffamily},
  lbl/.style={font=\scriptsize\sffamily\bfseries},
  arr/.style={->, thick},
  darr/.style={<->, thick},
  region/.style={draw, dashed, rounded corners=4pt, inner sep=6pt, thick},
  clawvm/.style={draw, rounded corners=4pt, inner sep=6pt, thick, fill=black!8},
]

\node[box, fill=white] (user) {User};

\node[box, fill=white, right=1.0cm of user, minimum width=2cm] (session) {Session\\Manager};
\node[box, fill=white, right=0.6cm of session, minimum width=2cm] (prompt) {Prompt\\Assembler};
\node[box, fill=white, right=0.6cm of prompt, minimum width=2cm] (toolmed) {Tool\\Mediator};

\node[sbox, fill=white, below=0.4cm of session] (compact) {Compaction};
\node[sbox, fill=white, right=0.3cm of compact] (prune) {Pruning};
\node[sbox, fill=white, right=0.3cm of prune] (flush) {Memory\\Flush};
\node[sbox, fill=white, right=0.3cm of flush] (reset) {Reset /\\Save};

\node[sbox, fill=black!12, below=1.0cm of compact, minimum width=1.6cm] (pgtbl) {Page\\Table};
\node[sbox, fill=black!12, right=0.3cm of pgtbl, minimum width=1.6cm] (selector) {Representation\\Selector};
\node[sbox, fill=black!12, right=0.3cm of selector, minimum width=1.6cm] (faults) {Fault\\Observer};
\node[sbox, fill=black!12, right=0.3cm of faults, minimum width=1.6cm] (wbjrnl) {Writeback\\Journal};

\node[box, fill=white, right=1.2cm of toolmed] (llm) {LLM};

\node[sbox, fill=white, below=1.1cm of llm] (tools) {Tool\\Backends};

\node[right=2.0cm of llm, anchor=north west, yshift=0.35cm] (cwanchor) {};

\node[draw, thick, rounded corners=2pt, minimum width=2.0cm, minimum height=4.75cm,
      anchor=north, fill=white, minimum width=2.1cm] (cw) at (cwanchor) {};
\node[lbl, fill=white, inner sep=1.5pt, anchor=south] at (cw.north) {Context Window};

\def\cwx{0}
\def\cww{1.65cm}

\node[draw, fill=gray!15, text width=\cww, align=center, minimum height=0.35cm,
      font=\tiny\sffamily, anchor=north] (cout) at ([yshift=-0.2cm]cw.north) {output reserve};

\node[draw, fill=gray!15, text width=\cww, align=center, minimum height=0.35cm,
      font=\tiny\sffamily, anchor=north] (csys) at ([yshift=-0.08cm]cout.south) {system prompt};

\node[draw, fill=gray!15, text width=\cww, align=center, minimum height=0.35cm,
      font=\tiny\sffamily, anchor=north] (ctool) at ([yshift=-0.08cm]csys.south) {tool schemas};

\node[draw, fill=green!20, text width=\cww, align=center, minimum height=0.65cm,
      font=\tiny\sffamily, anchor=north] (pfull) at ([yshift=-0.15cm]ctool.south) {page (full)};

\node[draw, fill=green!12, text width=\cww, align=center, minimum height=0.45cm,
      font=\tiny\sffamily, anchor=north] (pcomp) at ([yshift=-0.06cm]pfull.south) {page (compressed)};

\node[draw, fill=yellow!20, text width=\cww, align=center, minimum height=0.35cm,
      font=\tiny\sffamily, anchor=north] (pkey) at ([yshift=-0.06cm]pcomp.south) {page (structured)};

\node[draw, fill=orange!15, text width=\cww, align=center, minimum height=0.25cm,
      font=\tiny\sffamily, anchor=north] (pptr) at ([yshift=-0.06cm]pkey.south) {page (ptr)};

\node[draw, fill=white, text width=\cww, align=center, minimum height=0.55cm,
      font=\tiny\sffamily\itshape, text=black!40, anchor=north] (pfree) at ([yshift=-0.06cm]pptr.south) {free};

\draw[decorate, decoration={brace, amplitude=4pt}, thick]
  ([xshift=0.25cm]pfull.north east) -- ([xshift=0.25cm]pfree.south east)
  node[midway, right=5pt, font=\tiny\sffamily, align=left] {token\\budget};

\begin{scope}[on background layer]
  \node[region, fill=blue!4, fit=(session)(prompt)(toolmed)(compact)(prune)(flush)(reset)] (harness) {};
  \node[clawvm, fit=(pgtbl)(selector)(faults)(wbjrnl)] (clawvmbox) {};
\end{scope}

\node[lbl, fill=white, inner sep=1.5pt, anchor=south west] at (harness.north west) {Agent Harness};
\node[lbl, font=\scriptsize\sffamily\bfseries, fill=white, inner sep=1.5pt, anchor=south west] at ([xshift=-2pt]clawvmbox.north west) {\textsc{ClawVM} Layer};

\node[sbox, fill=white, below=1.2cm of pgtbl] (memfiles) {Memory\\Files};
\node[sbox, fill=white, right=0.3cm of memfiles] (transcripts) {Transcript\\Logs};
\node[sbox, fill=white, right=0.3cm of transcripts] (indexes) {Session\\Indexes};
\node[sbox, fill=white, right=0.3cm of indexes] (extmem) {External\\(QMD, Mem0)};

\begin{scope}[on background layer]
  \node[region, fill=orange!6, fit=(memfiles)(transcripts)(indexes)(extmem)] (durable) {};
\end{scope}
\node[lbl, fill=white, inner sep=1.5pt, anchor=south east] at ([yshift=2pt]durable.north east) {Durable Stores};

\draw[darr] (user) -- (session);
\draw[arr] (session) -- (prompt);
\draw[arr] (prompt) -- (toolmed);
\draw[darr] (toolmed) -- (llm);
\draw[darr] ([xshift=-4pt]toolmed.south east) -- ++(0,-0.35) -| (tools.north);
\draw[arr, densely dotted] (session) -- (compact);
\draw[arr] (pgtbl) -- (selector);
\draw[arr] (selector) -- (faults);
\draw[arr] (faults) -- (wbjrnl);
\path (prune.east) -- (flush.west) coordinate[midway] (pfgap);
\draw[arr, densely dashed] (selector.north) -- ++(0,0.6) -| (pfgap) -- (pfgap |- prompt.south);
\draw[arr, densely dashed] (compact.south) -- (pgtbl.north);
\draw[arr, densely dashed] (flush.south) -- ++(0,-0.5) -| (faults.north);
\draw[arr, densely dashed] (reset.south) -- ++(0,-0.5) -| (wbjrnl.north);
\draw[darr] (pgtbl.south) -- (memfiles.north);
\draw[darr] (wbjrnl.south) -- ++(0,-0.3) -| (transcripts.north);
\draw[arr, densely dashed] (extmem.north) -- ++(0,0.3) -| (faults.south);

\draw[arr, densely dashed] (llm.east) -- (cw.west |- llm.east) node[midway, above, font=\tiny\sffamily] {fills};

\node[font=\scriptsize\itshape, text=black!60, fill=white, inner sep=1pt, anchor=east] at ($(harness.west |- compact)+(-0.1,0)$) {(a)\,bootstrap loss};
\node[font=\scriptsize\itshape, text=black!60, fill=white, inner sep=1pt, anchor=west] at ($(harness.east |- reset)+(0.1,0)$) {(b)\,flush miss};
\node[font=\scriptsize\itshape, text=black!60, fill=white, inner sep=1pt, anchor=east] at ($(durable.west)+(-0.1,0)$) {(c)\,corrupt write};

\end{tikzpicture}
\caption{Architecture of a stateful tool-using agent with the \textsc{ClawVM} layer (shaded).
The agent harness manages sessions, assembles prompts, mediates tools, and emits lifecycle events (compaction, pruning, flush, reset).
\textsc{ClawVM} interposes at the harness level: the page table and representation selector feed prompt assembly, the fault observer instruments paging and lifecycle behavior, and the writeback journal enforces validated, non-destructive persistence at all lifecycle boundaries.
The context window (right) shows how pages at varying fidelity levels fill the token budget.
Letters mark failure points from field reports (Section~\ref{sec:background}): (a)~post-compaction bootstrap loss, (b)~flush miss on reset, (c)~destructive writeback to durable stores.}
\Description{Architecture diagram showing the ClawVM layer within a stateful tool-using agent harness, with labeled failure points at lifecycle boundaries.}
\label{fig:architecture}
\end{figure*}
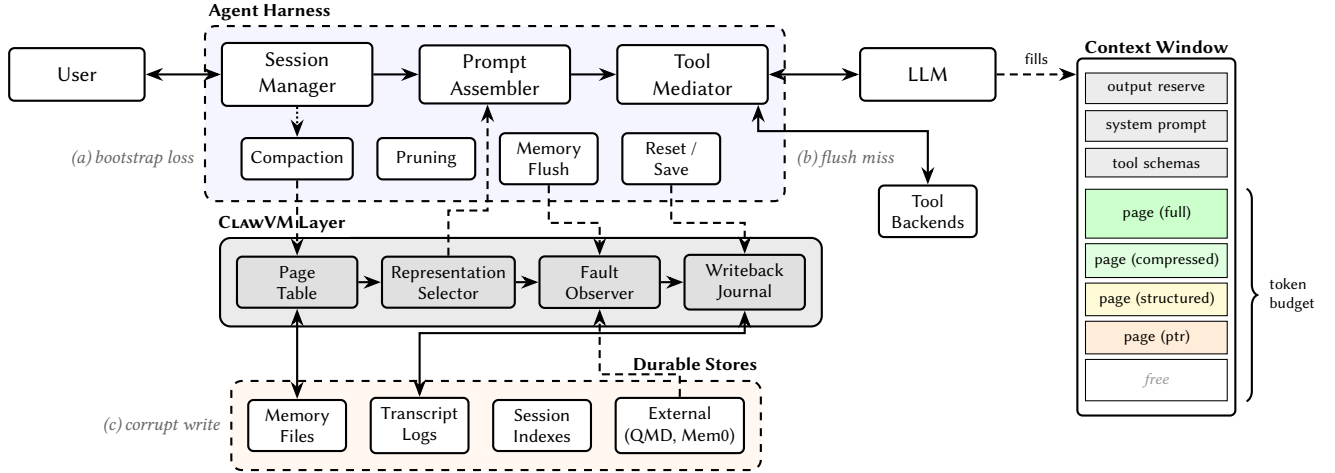

We present \textsc{ClawVM}, a harness-managed virtual memory layer that closes this gap (Figure~\ref{fig:architecture}).
\textsc{ClawVM} manages agent state as typed \emph{pages} that can be kept at full detail, compressed, reduced to structured fields, or shrunk to a pointer under token-budget pressure.
Each page carries a minimum-fidelity invariant (how far it may degrade before reclaiming space), and the harness enforces these invariants at every lifecycle boundary through staged, validated writeback.
When invariants are violated or state is lost, the system raises observable \emph{faults} that make the failure diagnosable and replayable.

Only the agent harness mediates between the model and its state.
Structurally, it is an OS kernel for agent state: it controls what stays resident and when state becomes durable.
Placing the memory contract here makes residency and durability explicit, deterministic, and replayable without retraining models or replacing retrieval backends.
Multi-resolution representations prevent compaction from being all-or-nothing: pages degrade gracefully, preserving enough to reconstruct.
Validated writeback prevents the destructive overwrites of today's free-form flush prompts.
Together, multi-resolution representations, validated writeback, and observable faults close three gaps that retrieval quality, compaction tuning, and external memory cannot: critical state must survive destruction, dirty state must be committed before it is lost, and recall failures must be diagnosable.

We implement a \textsc{ClawVM} prototype with a harness integration layer and evaluate the policy and fault model on OpenClaw-derived workloads with deterministic replays, comparing against retrieval-only and practitioner-configured compaction+retrieval baselines~\cite{openclaw:compaction,openclaw:config,openclaw:memory}.
We also validate the hook integration in a live harness across lifecycle regressions, deterministic replay suites, and trace-derived replays from real coding-agent sessions, following the evaluation methodology of recent memory benchmarks~\cite{amemgym2025,mem2actbench2026}.

Across four OpenClaw-derived workload families and six token-budget levels, \textsc{ClawVM} eliminates all policy-controllable faults (refetch, duplicate-tool, post-compaction bootstrap, and flush-miss) from 67.8 mean faults per workload-budget configuration (retrieval baseline) and 1.5 (practitioner-configured compaction+retrieval) to zero, and reduces paging instability by 77.4\% and 11.4\% respectively.
An offline oracle with future knowledge of demand confirms zero remaining headroom: the online policy already achieves the optimum fault count.
These results hold across all token budgets, 12 diverse real-session traces, and 30 task-level replays (100\% success vs.\ 76.7\% for the practitioner-configured baseline at the tightest budget), while adding median $<$\,50\,$\mu$s of policy-engine overhead per turn.

This paper makes the following contributions:
\begin{itemize}[leftmargin=1.5em]
  \item A virtual memory contract for agent state: typed pages with minimum-fidelity invariants and multi-resolution representations under a token budget.
  \item A fault model that makes memory-management decisions observable and replayable, with offline oracle analysis for tuning and regression.
  \item A staged writeback protocol with deterministic, non-destructive commit at every lifecycle boundary.
  \item A \textsc{ClawVM} prototype evaluated on synthetic, trace-derived, and adversarial workloads.
\end{itemize}

\section{Background and Motivation}
\label{sec:background}
Stateful tool-using agents operate under a long-running control plane, the \emph{agent harness}, that routes sessions, assembles each model call from conversational context and durable state, mediates tool invocation, and emits lifecycle events such as compaction, pruning, and reset~\cite{openclaw:context,openclaw:sessionmgmt}.
Because the harness manages both tiers, every prompt-assembly decision (what to include, at what fidelity, what to drop) is a page-replacement decision: the context window is physical memory, durable stores are disk.

Harnesses like OpenClaw provide practical building blocks: pruning low-value spans, retrieval via BM25 (term-frequency ranking) and vector search, compacting transcripts near the model's limit, and inserting a pre-compaction flush turn that asks the model to externalize durable notes~\cite{openclaw:context,openclaw:compaction,openclaw:memory}.
External layers (Mem0, Cognee, QMD) add auto-capture, hybrid recall, and graph-backed retrieval~\cite{mem0:openclaw,cognee:openclaw,qmd}.
These mechanisms improve recall but treat residency and durability as best-effort: nothing enforces what must stay resident, when state becomes durable, or how paging decisions can be audited.

The consequences are concrete.
Consider an agent 40~turns into a morning routine orchestrating email triage, calendar conflicts, and smart-home configuration, with loaded preferences (meeting priorities, lighting schedules, notification routing), a multi-step plan, and results from a dozen tool calls.
Compaction fires: the summary preserves the high-level goal but drops the plan's current step, routing rules, and evidence that conflicts were already resolved.
The agent re-queries calendars it already inspected (duplicate tool calls), routes a notification to a deprioritized channel (bootstrap loss), and restarts from step~1 of a plan it had completed through step~5.
On reset, 40~turns of accumulated decisions vanish; the runtime evicts dirty state without writing it back, and reload carries no guarantees.

These failures are not isolated.
Field reports document three recurring classes at lifecycle edges: \emph{residency failures} (compaction drops directives and tool outputs~\cite{openclaw:bootstrap-issues,openclaw:issue10524,openclaw:pr20267,practitioner:protocol-forgotten,practitioner:contextloss}), \emph{durability failures} (flushes bypassed on reset, destructive overwrites~\cite{openclaw:flush-bypass,openclaw:issue8185,openclaw:issue17034,openclaw:pr17041,openclaw:flush-stale,openclaw:writeback-corrupt}), and \emph{observability failures} (empty recall without reason codes~\cite{openclaw:silent-recall,openclaw:injection-hook}).
Practitioners trace these to three root causes: \textbf{capture is optional}, \textbf{recall is optional}, and \textbf{compaction is destructive}~\cite{openclaw:memory,mem0:blog,practitioner:bestpractices}.

Existing mitigations (tuned configs, external memory layers~\cite{openclaw:config,qmd,mem0:openclaw,cognee:openclaw,practitioner:compaction-fix}) cannot create an enforceable contract: the harness controls prompt assembly, tool mediation, and lifecycle events, yet exercises that control without an enforceable contract.

From these observations we derive six requirements, independent of model choice or retrieval backend:

\begin{itemize}[leftmargin=1.5em]
  \item \textbf{Invariants survive destruction.}
  The harness must restore instructions and constraints deterministically after compaction and reset.
  \item \textbf{Capture and recall are policy, not discretion.}
  For designated state, the harness drives capture and recall instead of relying on the model to flush or search.
  \item \textbf{Durability is lifecycle-complete.}
  The harness must commit dirty state at every boundary where the runtime would otherwise destroy the only copy.
  \item \textbf{Writeback is validated and non-destructive.}
  Updates must pass deterministic checks and use append/merge semantics; destructive overwrites must be rejectable.
  \item \textbf{Recall is observable.}
  The runtime must distinguish ``no match,'' ``denied,'' and ``backend error'' via reason codes.
  \item \textbf{Eviction is cost-aware.}
  Eviction decisions must account for the cost of re-running tool calls.
\end{itemize}

\noindent\textsc{ClawVM} is a harness-managed virtual memory layer that satisfies these six requirements.
The next section describes its design: typed pages, multi-resolution representations, an observable fault model, and a validated writeback protocol.

\section{Design}
\label{sec:design}

\textsc{ClawVM} addresses the six requirements from Section~\ref{sec:background} through three enforced choices at the harness level.
We first formalize the budget constraint, then describe each choice.

We model the agent runtime as an event stream over sessions.
Each event may trigger tool execution, memory lookup, and a model call.
After reserving tokens for output, system prompts, tool schemas, and a safety margin, the remaining capacity is the \emph{token budget} for paged state.
At each call, the memory manager selects a resident set that satisfies minimum-fidelity invariants within this budget and commits dirty state at lifecycle boundaries.
\textsc{ClawVM} makes three choices explicit and enforceable:
\emph{(i)} the unit of residency is typed pages,
\emph{(ii)} admissible degradations under budget pressure are multi-resolution representations, and
\emph{(iii)} lifecycle points at which state becomes durable use validated writeback.
Together these yield deterministic prompt assembly, controlled degradation instead of lossy summarization, observable faults, and lifecycle-complete persistence.

\paragraph{Pages as the unit of residency.}
\textsc{ClawVM} treats all assistant-relevant state as typed pages.
A page is a typed record with a stable identifier, scope, provenance, and a minimum-fidelity invariant.
Pages are the unit of selection, eviction, and writeback, turning best-effort prompting into enforceable behavior.
Scope (session-private vs.\ project-shared) and provenance (originating tool call or transcript span) are first-class metadata, required for observability and safe writeback.
Table~\ref{tab:pagetypes} lists six page types that cover failure-prone state, each with a minimum representation level (the four levels are defined below): \emph{Bootstrap/Policy} (system instructions and procedural directives, whose loss causes ``forgot its protocol'' failures), \emph{Constraint} (hard-pinned, never degraded below structured), \emph{Plan} (goal and current step while active), \emph{Preference} (scoped, may degrade to pointer), \emph{Evidence} (tool outputs with deterministically resolvable pointers), and \emph{Conversation Segment} (span identifiers and timestamp ranges).

\begin{table}[t]
\centering
\caption{Page classes, minimum-fidelity invariants, scope rules, and degradation paths in \textsc{ClawVM}.}
\label{tab:pagetypes}
\small
\setlength{\tabcolsep}{3pt}
\resizebox{\columnwidth}{!}{%
\begin{tabular}{L{1.8cm}L{2.2cm}L{0.8cm}L{2.5cm}}
\toprule
Type & Min.\ fidelity & Scope & Degradation path \\
\midrule
Bootstrap & Struct.\ after compact, reset & P/S & F{$\to$}St \\
Constraint & Hard-pinned at struct. & S & F{$\to$}St \\
Plan & Goal + step & S & F{$\to$}St{$\to$}Pt \\
Preference & Scope, prov. & P & F{$\to$}C{$\to$}St{$\to$}Pt \\
Evidence & Ptr resolves det.\ & S & F{$\to$}C{$\to$}St{$\to$}Pt \\
Conversation & Span ID + time & S & F{$\to$}C{$\to$}St{$\to$}Pt \\
\bottomrule
\end{tabular}}\\[2pt]
{\footnotesize F=full, C=compr., St=struct., Pt=ptr, P=proj., S=sess.}
\end{table}

\paragraph{Multi-resolution residency.}
Each page can be resident at one of four levels: \textit{full} (verbatim excerpt), \textit{compressed} (token-reduced text, e.g., LLMLingua-2~\cite{llmlingua2}), \textit{structured} (typed fields sufficient to satisfy invariants), and \textit{pointer} (a resolvable handle plus minimal metadata).
Under pressure, pages degrade along this chain while preserving invariants: constraints never drop below structured, and evidence pointers must remain resolvable.
Representations are generated at page creation time, not on demand under budget pressure: the harness extracts structured fields and computes token-reduced variants when a page is first ingested or updated.
The representation selector then chooses among pre-existing variants, so budget-pressure decisions involve only table lookups and token arithmetic, not runtime LLM calls or compression passes.
This prevents compaction from being all-or-nothing: structured and pointer representations preserve enough to reconstruct the full page on demand.

\paragraph{Fault model.}
Unlike OS page faults, which the kernel resolves transparently from disk, agent faults are silent: without instrumentation the harness cannot detect missing state, and recovery means regeneration.
\textsc{ClawVM} defines observable faults in two families, operationalizing the residency, durability, and observability failures from Section~\ref{sec:background}.
\emph{Working-set faults} capture missing state (residency failures): \emph{refetch faults} (a tool result re-retrieved after eviction), \emph{duplicate tool faults} (an equivalent tool call runs again because the result was evicted), \emph{pinned invariant misses} (hard-pinned page missing at prompt assembly), and \emph{post-compaction bootstrap faults} (required Bootstrap/Policy pages missing after compaction).
\emph{Durability faults} capture state lost at lifecycle boundaries (durability and observability failures): \emph{silent-recall faults} (lookup returns empty when the backend actually denied access or errored) and \emph{flush-miss faults} (dirty pages lost because the runtime destroyed the context before committing them).
These are \emph{policy-controllable} faults: a correct policy can prevent them.
Physical insufficiency (token budget too small to fit all pinned pages) and semantic errors (model produces factually wrong updates) are outside policy control and evaluated separately.
We also log \emph{duplicate-signature alerts} when canonical tool signatures repeat despite a resident result; these are workload signals, not policy failures.
We quantify instability with the \emph{thrash index}: the ratio of paging events to hits over the run (Appendix~\ref{app:design}), adapting the classical thrashing ratio~\cite{denning1968working}; high values signal a working-set/budget mismatch.

\paragraph{Selection policy.}
Prompt assembly is a multi-choice knapsack with hard constraints.
\textsc{ClawVM} uses a deterministic two-phase policy: \emph{Phase~1} installs all hard-pinned pages and minimum-required representations (surfacing invariant pressure if the minimum set cannot fit); \emph{Phase~2} applies greedy upgrades (pointer$\to$structured$\to$compressed$\to$full) by marginal utility per token until the budget is exhausted.
Utility combines pin status, membership, recency, scope, and recompute cost, with deterministic tie-breaks (Appendix~\ref{app:design}).
Listing~\ref{lst:assemble} gives the full algorithm.

\begin{lstlisting}[style=pseudocode,caption={Deterministic multi-resolution prompt assembly in \textsc{ClawVM}.},label={lst:assemble}]
function assemblePrompt(session, candidates, budget):
  prompt <- fixedRegion(session)
  prompt <- prompt + minimumRequired(candidates)
  remaining <- budget - tokens(prompt)

  upgrades <- allRepresentationUpgrades(candidates)
  sort upgrades by (deltaUtility(u) / deltaTokens(u)) descending

  for u in upgrades:
    if deltaTokens(u) <= remaining and respectsInvariants(u) and noConflict(u):
      applyUpgrade(prompt, u)
      remaining <- remaining - deltaTokens(u)

  return prompt
\end{lstlisting}

\paragraph{Replay oracle.}
To separate policy quality from budget insufficiency, \textsc{ClawVM} supports offline replay-oracle analysis: given a trace and the same budget, an oracle with bounded future knowledge picks representations that minimize faults.
The oracle's \emph{horizon} $h$ is the number of future demand turns visible; $h = \infty$ denotes unbounded lookahead.
The \emph{oracle gap} (online minus oracle fault counts) measures headroom versus unavoidable workload pressure.

\paragraph{Validated writeback protocol.}
\textsc{ClawVM} treats persistence as a three-phase lifecycle-aware transaction: \emph{structured staging} (the harness requests only typed, append/merge/set-with-version updates at lifecycle boundaries), \emph{deterministic validation} (each staged update is checked for schema correctness, provenance, scope, non-destructive semantics, and policy compliance), and \emph{scoped commit} (validated updates commit via deterministic merge rules at logged commit points).
Rejected updates remain in the journal with reason codes; full rules appear in Appendix~\ref{app:design}.

\section{Implementation}
\label{sec:implementation}
We implement a \textsc{ClawVM} prototype in six Python modules (\textasciitilde{}1,\!300 lines of non-comment code, zero external dependencies) targeting harnesses that expose session metadata, transcript logs, tool mediation, and lifecycle hooks~\cite{openclaw:sessionmgmt}.
Six lifecycle hooks map to engine operations; state is persisted via JSON snapshots between invocations.
Pluggable adapters normalize diverse sources (workspace files, transcript indexes, hybrid retrieval, QMD/Mem0/Cognee~\cite{qmd,mem0:openclaw,cognee:openclaw}) into \textsc{ClawVM} pages, keeping the contract retrieval-agnostic.
The six modules are: SessionPageTable, RepresentationSelector, WritebackJournal, FaultObserver, DecisionTrace (append-only JSON-lines audit log), and ClawVMEngine.

\section{Evaluation}
\label{sec:evaluation}

\paragraph{Setup.}
All experiments drive the \textsc{ClawVM} engine directly over pre-generated workload specifications, bypassing the harness hook layer to isolate selection and fault-detection logic.
Each workload defines pages (types, pin classes, token costs), per-turn demands, tool-call signatures, dirty-page sets, and lifecycle events; the engine runs the full per-turn pipeline and emits JSON-lines decision traces.
Token budgets range from 120 to 500 in six steps; all runs are deterministic.
Unless noted otherwise, aggregates average over 4~workloads \(\times\) 6~budgets \(= 24\) configurations.
Each RQ maps to the requirements from Section~\ref{sec:background}; fault types and workloads are chosen so every requirement is exercised.

\paragraph{Workloads.}
Four synthetic families exercise different stress patterns: \emph{evidence-heavy} (rotating tool signatures, periodic compaction), \emph{interruption-heavy} (parallel tasks with compaction and reset), \emph{lifecycle torture} (compaction every 3 turns, persistent dirty pages), and \emph{multi-session interference} (2 concurrent sessions, interleaved compaction).
Six lifecycle regression scenarios serve as a pass/fail gate (Table~\ref{tab:tier1}).

\paragraph{Baselines.}
We compare five policies (Table~\ref{tab:configs}): \emph{Retrieval} (no caching or writeback), \emph{Retrieval+Cache} (Retr.+Cache; adds artifact caching), \emph{Compaction-Hybrid} (Comp-Hybrid; prefetch + writeback at compaction but not reset, recency upgrades~\cite{openclaw:config}), \textsc{ClawVM} (full lifecycle completeness), and \emph{Oracle} with horizon $h\!=\!3$ (3-turn lookahead).
Comp-Hybrid represents best-case practitioner effort: its configuration (prefetch, compaction writeback, recency upgrades, pointer resolution) is distilled from OpenClaw documentation, community guides, and issue-tracker discussions~\cite{openclaw:config,practitioner:bestpractices,practitioner:compaction-fix}.
A real operator would need to discover and maintain each of these knobs independently.
Even so, Comp-Hybrid has structural gaps that no amount of configuration can close: it lacks reset writeback (dirty state lost on \texttt{/new}), hard auto-pinning (bootstrap pages unprotected), and utility-based upgrades (Table~\ref{tab:configs}).
No hyperparameter search was performed for any policy; all run under equal token budgets.

\begin{table}[t]
\centering
\caption{Policy configuration knobs (\cmark{} = enabled, -- = disabled).
Pin: auto-pin bootstrap/constraint pages.
Pref.: prefetch before demand.
WB-C/WB-R: writeback at compaction/reset.
Upgrade: Phase~2 scoring strategy (none, recency, utility, or oracle).
Resolve: follow pointer handles to reconstruct evidence content.}
\label{tab:configs}
\small
\setlength{\tabcolsep}{4pt}
\begin{tabular}{lcccccc}
\toprule
Policy & Pin & Pref. & WB-C & WB-R & Upgrade & Resolve \\
\midrule
Retrieval        & --    & --    & --    & --    & none & --    \\
Retr.+Cache      & --    & --    & --    & --    & none & \cmark \\
Comp-Hybrid      & --    & \cmark & \cmark & --    & recency & \cmark \\
\textbf{\textsc{ClawVM}} & \textbf{\cmark} & \textbf{\cmark} & \textbf{\cmark} & \textbf{\cmark} & \textbf{utility} & \textbf{\cmark} \\
Oracle ($h\!=\!3$) & \cmark & \cmark & \cmark & \cmark & oracle & \cmark \\
\bottomrule
\end{tabular}
\end{table}

\smallskip\noindent We organize the evaluation around three questions:
\begin{enumerate}[leftmargin=*, label=\textbf{RQ\arabic*}]
  \item How effectively does \textsc{ClawVM} eliminate policy-controllable faults?
  \item Which design features of \textsc{ClawVM} are necessary?
  \item Does \textsc{ClawVM} generalize beyond synthetic workloads, and at what cost?
\end{enumerate}

\subsection{RQ1: Fault Elimination}

We run all five policies (Table~\ref{tab:configs}) across 4 workloads and 6 token budgets (120--500), measuring explicit faults (refetch, duplicate-tool, bootstrap, flush-miss, silent-recall) and thrash (Section~\ref{sec:design}).
Six Tier-1 regression scenarios (Table~\ref{tab:tier1}) serve as a pass/fail gate.

\begin{table}[t]
\centering
\caption{Tier-1 lifecycle regression scenarios and gate assertions.}
\label{tab:tier1}
\small
\begin{tabular}{L{3.0cm}L{4.0cm}}
\toprule
Scenario & Assertion \\
\midrule
Post-compaction bootstrap & Bootstrap fault observable after compaction \\
Reset dirty-page flush miss & Flush-miss fault at reset boundary \\
Threshold jump race & Flush-miss fault from skipped pre-compaction commit \\
Silent recall visibility & Silent-recall fault from deny/error with reason code \\
Unsafe persistence rejection & Writeback rejected with \texttt{DESTRUCTIVE\_OP} reason \\
Evidence churn duplicate tool & Duplicate-tool fault from repeated equivalent call \\
\bottomrule
\end{tabular}
\end{table}

All six Tier-1 regression tests trigger the expected fault or rejection with the correct reason code, validating instrumentation and enforcement.
Table~\ref{tab:reductions} presents aggregate reductions across 24 configurations.
Against Comp-Hybrid, \textsc{ClawVM} reduces explicit faults by 100\% (1.5 mean to zero) and thrash by 11.4\%.
The 1.5 mean understates the gap: it averages over six budgets including loose ones (360--500) where all policies approach zero.
At the tightest budget (120 tokens), Comp-Hybrid incurs 26 faults across 4 workloads, including 7 bootstrap and 5 flush-miss, while \textsc{ClawVM} produces zero.
These are not tuning failures; they are structural: Comp-Hybrid lacks reset writeback and hard pinning, so no configuration change can prevent them.
Against retrieval, mean explicit faults drop from 67.8 to zero; thrash drops by 77.4\%.
Residual thrash comes from duplicate-signature alerts (inherent tool-signature reuse, not policy failures).
These fault classes map directly to field-reported symptoms: refetch and duplicate-tool faults produce redundant API calls~\cite{practitioner:contextloss}, bootstrap faults cause the agent to ``forget its protocol''~\cite{openclaw:bootstrap-issues,openclaw:issue10524,openclaw:pr20267,practitioner:protocol-forgotten}, and flush-miss faults mean decisions silently vanish on reset~\cite{openclaw:flush-bypass,openclaw:issue8185,openclaw:issue17034,openclaw:pr17041}.
Section~\ref{sec:limitations} discusses the remaining semantic-correctness gap.

\begin{table}[t]
\centering
\caption{Aggregate reductions relative to baselines, averaged over 4 workloads $\times$ 6 budgets (120--500 tokens).
\emph{Explicit faults}: policy-controllable events (refetch, duplicate-tool, bootstrap, flush-miss, silent-recall).
\emph{Thrash}: paging instability (faults + duplicate-signature alerts over hits); residual thrash at zero faults reflects workload repetition.}
\label{tab:reductions}
\small
\begin{tabular}{lrrrr}
\toprule
Baseline & Explicit & $\Delta$\% & Thrash & $\Delta$\% \\
\midrule
Retrieval        &  67.8 & $-$100\% & 3.993 & $-$77.4\% \\
Retr.+Cache      &  10.8 & $-$100\% & 1.642 & $-$45.2\% \\
Comp-Hybrid      &   1.5 & $-$100\% & 1.017 & $-$11.4\% \\
\midrule
\textbf{\textsc{ClawVM}} & \textbf{0.0} & \textbf{---} & \textbf{0.901} & \textbf{---} \\
Oracle ($h\!=\!3$)         &   0.0 &      0.0\% & 0.901 &   0.0\% \\
\bottomrule
\end{tabular}
\end{table}

\paragraph{Takeaway.} \textsc{ClawVM} eliminates all policy-controllable faults at every budget level, matching an offline oracle ($h\!=\!3$) on fault count.

\subsection{RQ2: Feature Necessity}

We assess which design features are necessary for fault elimination through subtractive and additive ablation.
We remove one feature at a time from full \textsc{ClawVM} at budget 180 (the tightest non-trivial budget) and measure explicit faults summed across all 4 workloads.

Table~\ref{tab:ablation} presents the subtractive ablation results.
Removing pointer resolution causes the most damage (126 faults across refetch and duplicate-tool), followed by writeback at compaction (20 flush-miss faults) and auto-pinning (9 bootstrap faults).
Removing upgrade scoring or prefetch yields zero faults; other features compensate, demonstrating redundancy and no single point of failure.
The additive ablation (Table~\ref{tab:additive}) confirms: pointer resolution alone eliminates 84\% of faults (271$\to$43), auto-pinning eliminates all bootstrap faults, and compaction writeback cuts flush-miss by 87\%.

\begin{table}[t]
\centering
\caption{Additive ablation at budget 180: explicit faults when enabling one feature on the bare baseline. Summed across 4 workloads.
Boot. = bootstrap, Flush = flush-miss, D+R = duplicate-tool + refetch.}
\label{tab:additive}
\small
\begin{tabular}{lrrrr}
\toprule
Variant & Boot. & Flush & D+R & Total \\
\midrule
baseline       &  20 &  23 & 228 &  271 \\
+pin           &   0 &  23 & 228 &  251 \\
+resolve       &  20 &  23 &   0 &   43 \\
+WB-C          &  20 &   3 & 228 &  251 \\
+WB-R          &  20 &  20 & 228 &  268 \\
+upgrade       &   7 &  23 &  42 &   72 \\
\textbf{\textsc{ClawVM}} & \textbf{0} & \textbf{0} & \textbf{0} & \textbf{0} \\
\bottomrule
\end{tabular}
\end{table}

\begin{table}[t]
\centering
\caption{Subtractive ablation at budget 180: explicit faults when removing one feature from full \textsc{ClawVM}. Summed across 4 workloads.
Boot. = bootstrap, Flush = flush-miss, D+R = duplicate-tool + refetch.}
\label{tab:ablation}
\small
\begin{tabular}{lrrrr}
\toprule
Variant & Boot. & Flush & D+R & Total \\
\midrule
\textbf{\textsc{ClawVM}} & \textbf{0} & \textbf{0} & \textbf{0} & \textbf{0} \\
$-$pin           & 9 & 0 & 0 & 9 \\
$-$resolve       & 0 & 0 & 126 & 126 \\
$-$WB-C          & 0 & 20 & 0 & 20 \\
$-$WB-R          & 0 & 3 & 0 & 3 \\
$-$upgrade       & 0 & 0 & 0 & 0 \\
$-$prefetch      & 0 & 0 & 0 & 0 \\
\bottomrule
\end{tabular}
\end{table}

\paragraph{LRU baseline.}
To test whether the utility-based knapsack is necessary or a simpler heuristic suffices, we add a least-recently-used (LRU) variant that keeps all structural features (auto-pinning, writeback, prefetch, pointer resolution) but replaces utility scoring with pure recency-ordered upgrades.
Across all 4 workloads and budgets 120--300, LRU achieves zero explicit faults and identical thrash to \textsc{ClawVM}.
This result reinforces the architectural claim: \textsc{ClawVM}'s fault elimination is robust by construction, guaranteed by the Phase~1 structural constraints (hard-pinning, lifecycle writeback, pointer resolution) regardless of the Phase~2 upgrade heuristic.
Utility scoring operates in the quality regime \emph{above} the fault-free floor, determining which pages are upgraded to higher-fidelity representations when budget permits.
We retain utility scoring by design because LRU makes strictly worse upgrade decisions in predictable situations: it spends budget upgrading recently touched but cheap-to-recompute evidence over stale but expensive-to-recompute results, and downgrades infrequently accessed preferences to pointer even when their base utility is high.
Under tight budgets with bursty access patterns, this means the model sees more pointers and fewer full representations for high-value state, degrading downstream output quality without triggering a fault.
Utility scoring accounts for recompute cost, pin class, and scope, preferring upgrades with the highest marginal value per token.
Evaluating this quality gap requires end-to-end task metrics with a live model, which we leave to future work; for the structural safety properties evaluated here, any reasonable upgrade heuristic suffices once the enforcement layer is in place.

\paragraph{Weight sensitivity.}
We sweep each of the four utility scoring weights (pin boost hard $\in[0.5,4.0]$, pin boost soft $\in[0.0,1.5]$, recency $\in[0.1,1.2]$, recompute cost $\in[0.0,1.2]$) independently across all 4 workloads at budget 180 (21 configurations, 84 runs).
All configurations produce zero explicit faults and identical thrash (0.901).
The weights are not brittle: any setting within the tested ranges achieves fault elimination, because the structural constraints prevent faults independent of upgrade ordering.
Weights will differentiate policies in prompt-quality evaluations, where the fidelity mix of resident pages affects downstream model behavior.

\paragraph{Takeaway.} Pointer resolution, auto-pinning, and lifecycle writeback are the critical features; fault elimination is robust to the choice of upgrade heuristic and weight parameterization.
Utility scoring provides a principled extension point for optimizing prompt quality above the fault-free floor.

\subsection{RQ3: Generalization and Cost}

We evaluate generalization and runtime cost using real-trace replay, task-granularity replay, adversarial stress tests, and overhead measurement.

We validate on 12 trace-derived replay workloads from real Claude Code sessions spanning coding, research, DevOps, trading, ML ops, web scraping, long-form writing, email writing, and product design (100-turn truncations; Appendix~\ref{app:real}).
Conversion is deterministic (tool signatures canonicalized by name and argument schema, dirty pages inferred from tool outputs, lifecycle events preserved from transcript metadata); no traces were used to tune policy parameters.
\textsc{ClawVM} produces zero explicit faults across all 12 traces; Retrieval produces a median of 51 faults (range 49--77).
Comp-Hybrid leaks exactly one flush-miss per trace---dirty state silently lost on every session because it lacks reset writeback.
At 200 turns the pattern holds: zero faults for \textsc{ClawVM}, while retrieval-only faults scale to a median of 83 (range 71--109; Appendix Table~\ref{tab:trace_scale}).
Sensitivity analysis (50--400 turns) confirms \textsc{ClawVM} maintains zero faults at all lengths, while retrieval-only faults grow near-linearly.

To move beyond per-turn aggregates, we construct 30 synthetic task workloads across four categories (coding, debugging, writing, ops; 5--14 turns each) and replay them at budgets 180 and 300.
A task ``succeeds'' if it completes with zero explicit faults (controlled replays, not live execution).
This criterion is structural (schema violations, missing pages, lost commits), independent of model output quality; human inspection confirmed zero-fault runs preserved state needed for correct completion.
Table~\ref{tab:task_level} shows the results: \textsc{ClawVM} achieves 100\% task success (zero explicit faults) at both budgets.
Comp-Hybrid drops to 76.7\% at budget 180; all 7 failures are bootstrap faults in debugging tasks where mid-task compaction evicts unprotected bootstrap pages (Table~\ref{tab:tasktype}).
At budget 300 the failures vanish---not because the policy improves, but because the larger budget avoids triggering compaction.
This brittleness is the core issue: Comp-Hybrid's success depends on having enough headroom to sidestep its structural gaps, and tighter budgets expose them.
The identical thrash across all rows (1.36) is a workload floor: residual thrash at zero faults reflects inherent tool-signature repetition, not policy differences.

\begin{table}[t]
\centering
\caption{Task success by type and policy at two budgets. All 7 Comp-Hybrid failures are bootstrap faults in debugging tasks.}
\label{tab:tasktype}
\small
\begin{tabular}{lrcccc}
\toprule
 & & \multicolumn{2}{c}{Comp-Hybrid} & \multicolumn{2}{c}{\textsc{ClawVM}} \\
\cmidrule(lr){3-4}\cmidrule(lr){5-6}
Type & $n$ & 180 & 300 & 180 & 300 \\
\midrule
coding    &  9 &  9/9  &  9/9  &  9/9  &  9/9  \\
debugging &  8 &  1/8  &  8/8  &  8/8  &  8/8  \\
writing   &  2 &  2/2  &  2/2  &  2/2  &  2/2  \\
ops       & 11 & 11/11 & 11/11 & 11/11 & 11/11 \\
\midrule
total     & 30 & 23/30 & 30/30 & 30/30 & 30/30 \\
\bottomrule
\end{tabular}
\end{table}

\begin{table}[t]
\centering
\caption{Task-granularity replay: 30 synthetic tasks at two budgets.
Success = zero explicit faults for the task.}
\label{tab:task_level}
\small
\setlength{\tabcolsep}{2pt}
\begin{tabular}{llrrrrr}
\toprule
Policy & Budget & Success & Faults & Tok/task & Thrash & $p_{50}$\,($\mu$s) \\
\midrule
Comp-Hybrid      & 180 &  76.7\% & 0.23 & 1607 & 1.36 & 170 \\
Comp-Hybrid      & 300 & 100.0\% & 0.00 & 2649 & 1.36 & 164 \\
\textbf{\textsc{ClawVM}} & \textbf{180} & \textbf{100.0\%} & \textbf{0.00} & \textbf{1605} & \textbf{1.36} & \textbf{167} \\
\textbf{\textsc{ClawVM}} & \textbf{300} & \textbf{100.0\%} & \textbf{0.00} & \textbf{2639} & \textbf{1.36} & \textbf{166} \\
\bottomrule
\end{tabular}
\end{table}

\paragraph{Live hook validation.}
Twenty single-session tasks against a production agent (Table~\ref{tab:livehook}) show 20/20 success for both policies with zero faults and comparable overhead (16.6\,s vs.\ 16.9\,s).
The two policies tie because single-session tasks do not trigger the lifecycle boundaries (compaction, reset) where Comp-Hybrid's structural gaps manifest.

We stress-test \textsc{ClawVM} with three adversarial scenarios (Table~\ref{tab:adversarial}):
\emph{budget starvation} (budget 40, three hard-pinned pages requiring 60 tokens total),
\emph{extreme churn} (50 unique evidence pages in 50 turns, budget 180), and
\emph{cascade reset} (9 resets in 30 turns with persistent dirty pages).
Under churn and cascade resets, \textsc{ClawVM} achieves zero faults where Retrieval incurs up to 298 and Comp-Hybrid up to 9.
Under starvation, all policies incur 10 identical pinned-invariant-miss faults: when the budget physically cannot fit all hard-pinned pages, no policy can help, and \textsc{ClawVM} surfaces this as a diagnosable failure rather than silent degradation.

\begin{table}[t]
\centering
\caption{Adversarial explicit faults by policy.
Boot. = bootstrap, Pin = pinned-invariant-miss, Flush = flush-miss, D+R = duplicate-tool + refetch.}
\label{tab:adversarial}
\small
\setlength{\tabcolsep}{3pt}
\begin{tabular}{llrrrrr}
\toprule
Scenario & Policy & Boot. & Pin & Flush & D+R & Total \\
\midrule
\multirow{3}{*}{starvation}
& Retrieval    &   0 &  10 &  0 &   0 &  10 \\
& Comp-Hybrid  &   0 &  10 &  0 &   0 &  10 \\
& \textbf{\textsc{ClawVM}} & \textbf{0} & \textbf{10} & \textbf{0} & \textbf{0} & \textbf{10} \\
\midrule
\multirow{3}{*}{churn}
& Retrieval    &   9 &   0 &  9 & 280 & 298 \\
& Comp-Hybrid  &   9 &   0 &  0 &   0 &   9 \\
& \textbf{\textsc{ClawVM}} & \textbf{0} & \textbf{0} & \textbf{0} & \textbf{0} & \textbf{0} \\
\midrule
\multirow{3}{*}{cascade}
& Retrieval    &   4 &   0 & 13 &  60 &  77 \\
& Comp-Hybrid  &   0 &   0 &  7 &   0 &   7 \\
& \textbf{\textsc{ClawVM}} & \textbf{0} & \textbf{0} & \textbf{0} & \textbf{0} & \textbf{0} \\
\bottomrule
\end{tabular}
\end{table}
The policy-engine decision procedure adds median 18--44\,$\mu$s per turn ($p_{95}$$<$60\,$\mu$s, worst-case 114\,$\mu$s), excluding model and tool latency, and $<$\,83\,KB peak memory.

\paragraph{Takeaway.} \textsc{ClawVM} generalizes to real traces, task-level workloads, and adversarial scenarios with zero policy-controllable faults and negligible overhead.

\paragraph{Threats to validity.}
Claims are scoped to structural lifecycle faults, not semantic correctness.
Deterministic replay isolates policy behavior but not online effects (model nondeterminism, tool latency).

\section{Related Work}
\label{sec:related}

\textsc{ClawVM}'s contribution is \emph{enforcement}: critical state survives lifecycle transitions, dirty state is committed before destruction, and paging decisions are observable.

\paragraph{OS abstractions and transactional agents.}
Several systems share overlapping abstractions but target different problems: AIOS~\cite{aios2024} provides a multi-agent kernel with scheduling and context management; MemOS~\cite{memos2025} unifies representation and evolution across heterogeneous memory types; SagaLLM~\cite{sagallm2025} adds Saga-style transaction guarantees; Text2Mem~\cite{text2mem2025} defines a typed memory operation language with schema-level invariants; and Memory~OS~\cite{memoryos2025} uses OS-inspired hierarchical storage with segmented page organization.
\textsc{ClawVM} composes enforcement primitives (lifecycle-complete commit, minimum-fidelity invariants, observable faults) into a contract at the harness level; it is a runtime layer that can wrap any of them.

\paragraph{Agent memory systems.}
CoALA~\cite{coala2023} and Generative Agents~\cite{generativeagents2023} define frameworks for what memory an agent should have; \textsc{ClawVM} addresses how the harness \emph{enforces} that memory survives lifecycle transitions.
MemGPT~\cite{memgpt2023} is the closest predecessor, explicitly framing context management as OS-style virtual memory with model-driven paging.
Memory-as-Action~\cite{memact2025} similarly lets the model drive paging decisions.
A-MEM~\cite{amem2025} takes a different approach, organizing long-term memory as a Zettelkasten-like network with dynamic note construction and linking rather than prompt-time paging.
These systems improve adaptivity but leave residency and writeback to model discretion; \textsc{ClawVM} moves both to the harness with replayable enforcement.
The two approaches are potentially composable: a MemGPT-style model could serve as the paging heuristic inside \textsc{ClawVM}'s enforcement layer, gaining lifecycle-complete writeback and observable faults while retaining adaptivity.
LLMLingua-2~\cite{llmlingua2} compresses prompts, RAGCache~\cite{ragcache2024} caches retrieval results, and production layers (Mem0~\cite{mem0:openclaw}, Cognee~\cite{cognee:openclaw}, QMD~\cite{qmd}) improve recall quality. All are orthogonal to enforcement and usable as \textsc{ClawVM} components.

\paragraph{Benchmarks.}
AgentBench~\cite{agentbench2023}, WebArena~\cite{webarena2023}, and OSWorld~\cite{osworld2024} measure end-to-end task success but do not instrument paging or lifecycle failures.
LoCoMo~\cite{locomo2024}, LongMemEval~\cite{longmemeval2024}, AMemGym~\cite{amemgym2025}, Mem2ActBench~\cite{mem2actbench2026}, and MemoryAgentBench~\cite{memoryagentbench2026} benchmark model memory ability across retrieval, learning, and forgetting.
\textsc{ClawVM} evaluates enforcement at the runtime level: lifecycle faults, commit correctness, and replayable selection.

\section{Limitations}
\label{sec:limitations}

\balance
We identify three limitations that bound the scope of our claims.

\paragraph{Semantic correctness is out of scope.}
\textsc{ClawVM} validates schema, provenance, scope, and non-destructiveness, but does not verify the semantic truth of model-generated updates.
A model can produce a well-formed, properly scoped update that is factually wrong.
Semantic verification could be layered on via domain-specific validation predicates on the writeback journal.

\paragraph{Replay assumes simulatable tools.}
The replay engine models tool calls by canonical signatures without executing real tools.
Extending to non-deterministic services requires recording and replaying actual tool outputs, which the trace schema supports but the current prototype does not exercise.

\paragraph{External validity.}
Results cover replayed demand streams, 12 converted real-session traces, 30 synthetic task-level workloads, and 20 live single-session tasks.
The live experiment confirms hook correctness but exercises only single-session execution; cross-session lifecycle edges (compaction, reset) remain replay-validated only.
Multi-session online deployment with user-visible task-success metrics is future work.

\section{Conclusion}
\label{sec:conclusion}
Recurring memory failures in stateful tool-using agents are symptoms of a missing contract between the harness and the state it manages.
\textsc{ClawVM} provides the missing contract: typed pages, minimum-fidelity invariants, multi-resolution representations, and validated writeback at every lifecycle boundary.
Across synthetic, real-trace, and adversarial workloads at all budget levels, \textsc{ClawVM} eliminates all policy-controllable faults with negligible overhead; an offline oracle confirms zero remaining headroom on fault count.
Fault elimination is robust by construction: an LRU baseline with the same structural features achieves identical fault counts, and utility scoring weights are insensitive across wide ranges, confirming that safety comes from the enforcement layer rather than heuristic tuning.
The fault model and replay oracle are independently useful for evaluating alternative policies.
Future work includes end-to-end evaluation with live models, multi-session deployment, and extension to multi-agent orchestrators.

\paragraph{Artifact availability.} All source code, data, and scripts used in this paper are publicly available at \url{https://github.com/mpi-dsg/clawvm}.

\let\oldthebibliography\thebibliography
\renewcommand{\thebibliography}[1]{\oldthebibliography{#1}\small\setlength{\itemsep}{0pt}\setlength{\parsep}{0pt}\setlength{\parskip}{0pt}}
\bibliographystyle{ACM-Reference-Format}
\bibliography{main}

\clearpage
\appendix

\section{Design Details}
\label{app:design}

This appendix collects formal definitions behind the design choices in Section~\ref{sec:design}, provided for reproducibility and implementation reference.

\paragraph{Utility scoring.}
\label{app:utility}
The greedy upgrade order in Listing~\ref{lst:assemble} depends on a per-page utility function.
For page \(p\), representation \(r\), and turn \(t\), the utility is:
\[
\begin{aligned}
U_t(p,r)=&\;w_{\text{pin}}I_{\text{pin}}(p)+w_{\text{boot}}I_{\text{boot}}(p)+w_{\text{plan}}I_{\text{plan}}(p,t)\\
&+w_{\text{rec}}R_t(p)+w_{\text{scope}}S(p)+w_{\text{rc}}C(p,r),
\end{aligned}
\]
where \(I_{\text{pin}}\), \(I_{\text{boot}}\), \(I_{\text{plan}}\) are indicator functions for pin status, bootstrap/policy membership, and active-plan membership; \(R_t\) is recency; \(S\) is scope utility; and \(C\) is recompute cost.
Each candidate upgrade \(u\) is ranked by \(s(u)=\Delta U_t(u)/\Delta\mathrm{tokens}(u)\), with deterministic tie-breaks by \((\texttt{page\_id},\texttt{rep\_level})\).
In our experiments, the \textsc{ClawVM} score uses pin boost \((2.0\ \text{hard}, 0.6\ \text{soft})\), recency weight \(0.6\), and recompute-cost weight \(0.4\); the recency baseline uses \((0.9, 0.1)\). Oracle variants also weight each page by \(2.2\times\) the number of future demands remaining in the trace.

\paragraph{Thrash index.}
\label{app:thrash}
Section~\ref{sec:design} defines the thrash index informally; the precise formulation is:
\[
\mathrm{thrash}=\frac{F}{H+1},
\]
where \(F\) counts paging events (explicit faults plus duplicate-signature alerts) and \(H\) counts hits over the entire run; \(+1\) prevents division by zero.
Because duplicate-signature alerts reflect inherent workload repetition (re-issued tool signatures when the result is already resident), thrash can be non-zero even when all policy-controllable faults are eliminated.

\paragraph{Writeback protocol details.}
\label{app:writeback}
Section~\ref{sec:design} summarizes the three-phase writeback protocol; this appendix provides the full specification.

\emph{Phase~1 (structured staging):} at eviction/downgrade boundaries or lifecycle hooks (flush/reset/compaction), the harness requests structured updates only: tuples of (field, op, value, scope, evidence\_ref) where ops are append/merge/set-with-version. These are appended to a \texttt{WritebackJournal} but not yet applied.

\emph{Phase~2 (deterministic validation):} each staged update is checked against five invariants: (1)~schema correctness (field exists, types check, size limits hold), (2)~provenance (evidence\_ref resolves; dangling provenance rejected), (3)~scope (requested scope allowed for the active principal/session), (4)~non-destructive semantics (no overwrite of previously committed content), and (5)~policy/safety (no violation of hard-pinned constraints).
Rejected updates remain in the journal with reason codes (e.g., \texttt{SCOPE\_DENIED}, \texttt{DESTRUCTIVE\_OP}).

\emph{Phase~3 (scoped commit):} validated updates commit into the appropriate scoped store using deterministic merge rules at explicit, logged commit points.

\paragraph{Non-destructiveness rule.}
For each staged update targeting key \(k\), \textsc{ClawVM} checks the journal's last committed version \(v_k\).
Append/merge operations are valid only if they preserve prior committed entries for \(k\).
Set-with-version is valid only when the staged version equals \(v_k\); otherwise the update is rejected as \texttt{DESTRUCTIVE\_OP}.

\section{Supplementary Results}
\label{app:results}

This appendix provides supplementary evaluation tables behind the claims in Section~\ref{sec:evaluation}.

\paragraph{Real-trace validation and scaling.}
\label{app:real}
To test generalization beyond synthetic workloads, we converted 12 diverse real Claude Code session transcripts (coding, research, DevOps, trading, ML ops, web scraping, long-form writing, email writing, product design) into replay workloads.
Table~\ref{tab:trace_scale} shows aggregate fault counts at 100 and 200 turns, demonstrating how each policy scales with session length.
\textsc{ClawVM} maintains zero explicit faults at both lengths.
Comp-Hybrid leaks exactly one flush-miss per trace at 100 turns; at 200 turns the flush-miss no longer appears in the median, though the structural gap (missing reset writeback) remains.
Retrieval-only faults grow near-linearly, from a median of 51 at 100 turns to 83 at 200 turns.

\begin{table}[h]
\centering
\caption{Aggregate explicit faults across 12 real traces at two session lengths (budget 300). Median and range reported.}
\label{tab:trace_scale}
\small
\begin{tabular}{llrr}
\toprule
Policy & Turns & Median (range) & Thrash \\
\midrule
Retrieval      & 100 & 51 (49--77) & 0.677 \\
Retrieval      & 200 & 83 (71--109) & 0.527 \\
Comp-Hybrid    & 100 &  1 (1--1) & 0.047 \\
Comp-Hybrid    & 200 &  0 (0--0) & 0.029 \\
\textbf{\textsc{ClawVM}} & \textbf{100} & \textbf{0 (0--0)} & \textbf{0.032} \\
\textbf{\textsc{ClawVM}} & \textbf{200} & \textbf{0 (0--0)} & \textbf{0.032} \\
\bottomrule
\end{tabular}
\end{table}

\paragraph{Live hook validation.}
\label{app:livehook}
Table~\ref{tab:livehook} summarizes the live single-session validation.

\begin{table}[h]
\centering
\caption{Live hook validation: 20 single-session tasks with a production agent. Both policies achieve identical success with comparable overhead.}
\label{tab:livehook}
\small
\begin{tabular}{lrrr}
\toprule
Policy & Success & Faults & Mean time (s) \\
\midrule
Comp-Hybrid & 20/20 & 0 & 16.9 \\
\textsc{ClawVM} & 20/20 & 0 & 16.6 \\
\bottomrule
\end{tabular}
\end{table}

\end{document}